\documentclass[sigconf, authorversion]{acmart}

\AtBeginDocument{%
  \providecommand\BibTeX{{%
    \normalfont B\kern-0.5em{\scshape i\kern-0.25em b}\kern-0.8em\TeX}}}

\copyrightyear{2023}
\acmYear{2023}
\setcopyright{acmlicensed}\acmConference[FAccT '23]{2023 ACM Conference on Fairness, Accountability, and Transparency}{June 12--15, 2023}{Chicago, IL, USA}
\acmBooktitle{2023 ACM Conference on Fairness, Accountability, and Transparency (FAccT '23), June 12--15, 2023, Chicago, IL, USA}
\acmPrice{15.00}
\acmDOI{10.1145/3593013.3593996}
\acmISBN{979-8-4007-0192-4/23/06}

\begin{document}

%%
%% The "title" command has an optional parameter,
%% allowing the author to define a "short title" to be used in page headers.
\title{Making Intelligence: Ethical Values in IQ and ML Benchmarks}

% LIST SUBTITLE OPTIONS 
% 
%
% 
%  
%%
%% The "author" command and its associated commands are used to define
%% the authors and their affiliations.
%% Of note is the shared affiliation of the first two authors, and the
%% "authornote" and "authornotemark" commands
%% used to denote shared contribution to the research.
 \author{Borhane Blili-Hamelin}
 \authornote{Both authors contributed equally to this research.}
 \email{borhane.blilihamelin@gmail.com}
 \affiliation{
  \institution{AI Risk and Vulnerability Alliance}
   %\streetaddress{P.O. Box 1212}
  \city{Brooklyn}
  \state{New York}
  \country{USA}
  \postcode{11215}}
 \orcid{0000-0002-9573-3332}
 \author{Leif Hancox-Li}
 \authornotemark[1]
 \email{leif.hancox-li@capitalone.com}
 \affiliation{
\institution{Capital One}
\streetaddress{1600 Capital One Drive}
 \city{McLean}
 \state{Virginia}
 \postcode{22102}
 \country{USA}
 }
 \orcid{0009-0006-1355-9104}
%\affiliation{%
 % \institution{Institute for Clarity in Documentation}
  %\streetaddress{P.O. Box 1212}
  %\city{Dublin}
  %\state{Ohio}
  %\country{USA}
  %\postcode{43017-6221}
%}

%%
%% By default, the full list of authors will be used in the page
%% headers. Often, this list is too long, and will overlap
%% other information printed in the page headers. This command allows
%% the author to define a more concise list
%% of authors' names for this purpose.

%Commenting out for anonymous review
%\renewcommand{\shortauthors}{Blili-Hamelin and Hancox-Li}

%%
%% The abstract is a short summary of the work to be presented in the
%% article.
\begin{abstract}
 In recent years, ML researchers have wrestled with defining and improving machine learning (ML) benchmarks and datasets. In parallel, some have trained a critical lens on the ethics of dataset creation and ML research. In this position paper, we highlight the entanglement of ethics with seemingly ``technical'' or ``scientific'' decisions about the design of ML benchmarks. Our starting point is the existence of multiple overlooked structural similarities between human intelligence benchmarks and ML benchmarks. Both types of benchmarks set standards for describing, evaluating, and comparing performance on tasks relevant to intelligence---standards that many scholars of human intelligence have long recognized as value-laden. We use perspectives from feminist philosophy of science on IQ benchmarks and thick concepts in social science to argue that values need to be considered and documented when creating ML benchmarks. It is neither possible nor desirable to avoid this choice by creating value-neutral benchmarks. Finally, we outline practical recommendations for ML benchmark research ethics and ethics review.
\end{abstract}

%%
%% The code below is generated by the tool at http://dl.acm.org/ccs.cfm.
%% Please copy and paste the code instead of the example below.
%%


%%
%% Keywords. The author(s) should pick words that accurately describe
%% the work being presented. Separate the keywords with commas.

%% A "teaser" image appears between the author and affiliation
%% information and the body of the document, and typically spans the
%% page.
%\begin{teaserfigure}
 % \includegraphics[width=\textwidth]{sampleteaser}
 % \caption{Seattle Mariners at Spring Training, 2010.}
 % \Description{Enjoying the baseball game from the third-base
 % seats. Ichiro Suzuki preparing to bat.}
 % \label{fig:teaser}
% \end{teaserfigure}

%\received{6 February 2023}
%\received[revised]{12 March 2009}
%\received[accepted]{5 June 2009}

%%
%% This command processes the author and affiliation and title
%% information and builds the first part of the formatted document.
\maketitle

\section{Introduction}
As machine learning (ML) grows as an academic discipline and in social impact, research communities are paying increasing attention to the ethical risks of their work \cite{srikumar_advancing_2022, partnership_on_ai_managing_2021, prunkl_institutionalizing_2021, editorial_how_2021, sturdee_consequences_2021, bender_naacl_2021, lovelace_ethics_2022, lovelace_looking_2022, bengio_provisional_2022, samy_bengio_announcing_2023}, and to how they measure the success of their work through benchmarks. As an example, NeurIPS recently created a benchmarks and datasets track, which goes through an ethics review process similar to the main conference track, in which authors are asked to think about potential uses of their work and their treatment of human subjects.\footnote{For a helpful overview of  the multiple developments in AI research ethics that began in 2020, see \citet{srikumar_advancing_2022}.} In this position paper, we argue that the ethical risks of ML benchmarks go deeper, cutting to the scientific core of how benchmarks are imagined to be scientifically valid. Specifically, we draw lessons from the history of human intelligence measurement and apply them to the ML case. Similarly to IQ tests, ML benchmarks involve ethical risks that trouble the line between technical or scientific concerns (such as construct validity and generality), and ethical concerns (such as justice, respect for persons, autonomy, etc.).

Our paper makes two contributions. First, we provide a conceptual framework for examining the junctures at which seemingly purely technical or scientific decisions about ML benchmarks are and should be informed by ethical considerations. Second, we highlight  overlooked similarities between IQ and ML benchmarks, and argue that these allow ML benchmark researchers to learn some lessons from the history of IQ testing. IQ and ML benchmarks both set standards for quantitative description, evaluation, and comparison on tasks relevant to intelligence. Both involve the specification of standard tasks for comparison of performance. Moreover, both involve quantitative metrics of success on those tasks, sometimes followed by a step of weighting tasks to calculate overall rankings across different tasks \cite{wechsler_wechsler_2008, wang2018glue}.\footnote{By contrast, typical comparisons between ML and human intelligence tend to focus on similarities and differences between human cognitive abilities and the abilities of ML systems \cite{canaan2019leveling, marcus2019rebooting}. Likewise, attempts have been made at building datasets that enable measuring ML model performance on human IQ tests \cite{liu_how_2019}. This question of similarities and differences between human and ML abilities will not be the focus of our paper.} We contend that the ML benchmark community stands to benefit from paying close attention to these overlooked structural similarities. Specifically, we show how insights from feminist philosophy of science scholarship on IQ research and on thick concepts in social science helps to anticipate the following areas of ethical risk within technical decisions about ML benchmarks:\footnote{From our perspective as outsiders, the field of human intelligence research does not currently look like a welcoming home for the insights of its antiracist, feminist, and decolonial critics. We see this paper as a celebration of the lasting importance of their insights, even beyond the context of the field of human intelligence research.}

%\footnote{Our approach takes on the burden of arguing that each of the points we make about benchmarks apply because of features of ML benchmarks, not because benchmarks are similar to IQ. Rather, we use the IQ case to highlight how difficulties in defining IQ benchmarks are due to structural features in the problem that also exist in ML. We then independently argue that these same structural features of ML benchmarks lead to similar difficulties as those encountered in IQ.} 
\begin{enumerate}
    \item The ethical risks that come with task selection---selecting which tasks matter enough to be included in a benchmark. We discuss the IQ case in Section \ref{crosscultural} and the ML case in Section \ref{task-choice}.
    \item The ethical risks that come with choosing standards of construct validity, and with prioritizing specific forms of validity over others. We discuss the IQ case in Section \ref{validity} and the ML case in Section \ref{ml-validity}.
    %ML has chosen a form of internal validity (cross-validation) as its main method of measuring model performance. At the same time, it has largely neglected measures of external validity typically used in the social sciences. Through the argument for inductive risk (Section \ref{validity}), we argue that the choice of standards of validity should be informed by ethical values, and should be emphasized in ethics review of ML benchmark papers. Lack of external validity is a common failure mode for ML systems (Section \ref{ml-validity}). 
    \item The ethical risks that come with path dependence, where benchmarks and particular types of models or practices reinforce each other in a positive feedback loop. We discuss the IQ case in Section \ref{self} and the ML case in Section \ref{path-dependency}.
    
\end{enumerate}

After arguing for these claims, we outline (Section \ref{practical}) practical recommendations towards accounting for these areas of risk in ML research and applications. We conclude (Section \ref{good-science}) with a reflection on what good science means in the context of ML benchmarks. 

\section{Background}

 Since its inception, the FAccT community has investigated questions related to the ethical impacts of ML. At the same time, conferences like NeurIPS have provided guidance on how ethics review for papers should pay close attention to issues like the potential uses of ML and the treatment of human subjects \cite{prunkl_institutionalizing_2021, luccioni_ethical_2021, bengio_provisional_2022, lovelace_looking_2022, lovelace_ethics_2022, samy_bengio_announcing_2023}. These, however, are risks that are shared by all ML research. In this paper, we are interested in ethical risks that set benchmarks apart from other areas of ML research. We argue that much greater attention needs to be paid to ethical risks within the scientific and technical core of how benchmarks are designed. 

Following \citet{raji2021ai}, we understand ML benchmarks to involve a combination of dataset(s) and metrics, where the metrics attempt to capture model performance on task(s). Benchmarks influence practices in the ML research community and in the design of practical ML applications, where benchmarks are used to compare the performance of ML methods and models, sometimes through public leaderboards. Influential benchmarks shape research agendas on the specific ML tasks they center. They help determine the dominant paradigms of ML research and applications.\footnote{See \citet{dotan_value-laden_2020} for a helpful look at the values implicit in ImageNet's influence over the rise of deep learning as a dominant paradigm in ML research.} The practice of adopting benchmarks for comparison of model performance opens the door to areas of ethical risk that require extra attention from the benchmark research community. In our view, this is also the area where lessons from the case of \emph{human intelligence  research} are especially helpful. 

The areas of ethical risk we examine place our work at the intersection of two strands of scholarship on the methodology of ML research. Recent research on ML benchmarks and datasets has interrogated practices like the prioritization of benchmarks on tasks believed to be indicators of progress on general-purpose ability, the prioritization of internal validity over external validity, and SOTA-chasing on a handful of influential benchmarks \cite{raji2021ai, liao_are_2021, bowman_what_2021, dehghani_benchmark_2021}. These three tendencies overlap with the areas of ethical risk we identify: around task selection, standards of validity, and path dependence.

Our paper also intersects with a growing body of research showing how technical or scientific priorities in ML research turn out to be \emph{value-laden}: that is, dependent on social, political, and/or ethical values \cite{dotan_value-laden_2020, birhane_values_2022, leif_2022, scheuerman2021datasets, hutchinson_evaluation_2022}. \footnote{For instance, \citet{birhane_values_2022} argue that the most mentioned priorities in highly cited NeurIPS and ICML only appear purely scientific or technical on a superficial analysis. They argue that upon closer scrutiny, values like ``performance, generalization, building on past work, quantitative evidence, efficiency, and novelty'' tend to support the centralization of power in ML research.} There is a small but growing body of work on values in datasets \cite{scheuerman2021datasets, denton_bringing_2020, denton2021genealogy, mathur_disordering_2022}. By contrast, the value-laden aspects of ML benchmarks---understood as combinations of datasets and metrics---have received much less attention. One example is \citet{lacroix_metaethical_2022}'s argument against the idea that benchmarks are an appropriate tool for determining AI systems are `ethical’---i.e. whether decisions made by AI models are ``morally `correct'''. Another is \citet{bommasani_evaluation_2022}'s argument that ML evaluation needs to be understood as a political force for change: evaluation ``succeeds when it achieves the desired change in the ﬁeld''.

In this paper, we argue against the temptation to separate the obviously ethical or political aspects of benchmarks from their technical features. Having in mind the context of ethics review, we focus on  \emph{ethical values} in a broad sense: considerations around what actions are more or less worth pursuing, or what outcomes are more or less worth bringing about. Our analysis is compatible with understanding the relevant values as social, political, or moral.

\section{Values in Human Intelligence Research}
\label{human-thick}

What roles do ethical values play in measuring human intelligence? In this section, we examine this question with an eye to identifying helpful lessons for the case of ML benchmarks (Section \ref{ml-values}). This section also serves as an introduction to key philosophy of science concepts about the place of ethical values in scientific research: thick concepts in social science, the practical consequences of research, and the argument from inductive risk.

Our perspective is deeply informed by feminist philosophy of science scholarship on the value neutrality of science. (See Appendix \ref{value-neutrality} for more context.) We especially draw on Elizabeth Anderson's investigation of how antiracist, feminist research on IQ challenges strong conceptions of the value-neutrality of science, while---as we see in Section \ref{good-science} and \ref{value-neutrality}---preserving an important place for a limited form of value neutrality.\footnote{For an account of general parallels between the value-laden history of IQ research and AI, see \citet{cave_problem_2020}.} This section reconsiders and bolsters Anderson's key arguments, in order to later show how they transfer over to ML.

In the context of intelligence measurement, ``intelligence'' is used in a gradable sense: as picking out something that comes in degrees. Measurement of human intelligence is concerned with abilities that humans can have more or less of \citep{ritchie_intelligence_2015, cave_problem_2020}. ``Intelligence'' sometimes gets used in a categorical sense, as picking out a difference in kind: e.g. ``are there intelligent life forms on other planets''? Or in the phrase, ``machine learning is a type of artificial intelligence''. In this paper, we focus on the gradable sense of intelligence.

To set the stage, we introduce \emph{thick concepts} as a tool to help us understand the ethical challenges that are particular to measuring human intelligence.

\subsection{Intelligence is a Thick Concept}
%\label{human-thick}

Intelligence (in the gradable sense) is a thick evaluative concept \cite{anderson_situated_2002, cave_problem_2020}. Thick concepts blur the line between evaluation and description. They convey content both about how we evaluate the world (approval or disapproval, praise or blame, success or failure, etc.) and about features of the world that seem independent of our evaluations \cite{kirchin_introduction_2013, kirchin_thick_2017}. 

For instance, calling someone ``open-minded'' describes empirical features of how they tend to act across situations---and perhaps about their personality \cite{sep-thick-ethical-concepts}. But open-mindedness is also a term of \emph{epistemic praise}: expressing \emph{values, standards, or ideals} having to do with how we should form and justify beliefs, how we should respond to evidence or uncertainty, and when it is appropriate or praiseworthy to change one's mind---say, when presented with new evidence. 

As another example, consider the World Health Organization's definition of health as "a state of complete physical, mental, and social well-being and not merely the absence of disease or infirmity" \cite{world_health_organization_basic_2020}. Health is also a thick concept: it describes empirical features of organisms and evaluates the desirability or value of the condition of an organism (well-being, illness, etc.)

Intelligence is a thick concept, in the sense that it is a term of praise that also purports to describe empirical facts about people's abilities \citep{anderson_situated_2002, cave_problem_2020}. Recognizing this feature of the concept of intelligence is helpful to identifying otherwise overlooked areas where ethical values and risks play a role in intelligence research. In Section \ref{ml-values}, we will also see how it helps bring out the structural parallels between IQ and ML benchmarks.

We take the value neutrality of science to be at issue when ethical, social, or political values are used to make scientific decisions (see \ref{value-neutrality}). Many scientists recognize that the value of particular research projects or even areas of study can sometimes be outweighed by ethical values. This is why we have regulations about experiments involving humans and animals. However, in the case of measuring human and machine intelligence, values are embedded in a deeper way: in how we define the measures themselves (\ref{crosscultural}), in our selection of standards of validity (\ref{validity}), and through path dependence (\ref{self}).

\subsection{Values determine what counts as intelligence}
\label{crosscultural}

Values enter into defining the boundaries of the objects of intelligence research. Intelligence research is concerned with abilities. But what are their boundaries? Where do the abilities begin and end? What marks an ability as relevant or irrelevant to intelligence? Whose abilities is intelligence research about?

Consider questions about the relationship between the objects of intelligence research and cultural boundaries. Are the findings of intelligence research culturally specific? \citet{warne_spearmans_2019} argue that this concern is heightened by the fact that definitions of intelligence are not only variable across cultures but within them. Even within specific cultures, there is a lack of expert consensus on what generally falls under the term ``intelligence'', and on the specific abilities that matter to intelligence.

Following Anderson, we take boundary problems to come hand in hand with the fact that intelligence is a thick evaluative concept \cite{anderson_situated_2002}. There are cross-cultural variations and there is a lack of expert consensus on what falls under ``intelligence'' because ethical values and interests play a central role in determining both what empirical phenomena fall under the concept and the theoretical content of the concept. 

This problem is shared with research on topics like \emph{health} \cite{anderson_situated_2002} and \emph{well-being} \cite{alexandrova_democratising_2022}. Social science disciplines that inherit their subject matter from \emph{thick concepts} usually face problems with separating the definition of the boundaries of their topic from \emph{ethical} values. 

As Alexandrova and Fabian argue, a common strategy for sidestepping boundary problems with thick concepts in the social sciences is to attempt to convert the thick concepts into technical terms \cite{alexandrova_democratising_2022}. This strategy remains recently favored by some human intelligence researchers. For the sake of securing cross-culturally invariable boundaries for the object of intelligence research, \citet{warne_spearmans_2019} propose that researchers should focus on an object whose boundaries are simply a matter of ``statistical observation'': Spearman's \emph{g}.

 What researchers directly observe in cognitive tests is performance on very specific tasks \cite{spearman_general_1904, carroll_1993, deary_125_2012, deary_intelligence_2020}. For instance, one test contains up to 15 different tasks:  word similarity, vocabulary, visual puzzles, symbol search, digit span, comprehension, etc. \cite{wechsler_wechsler_2008, deary_intelligence_2020}. Early on, the field struggled with empirically studying mental abilities beyond performance on very specific tasks. Spearman is credited with realizing that these problems can be sidestepped through a technical procedure: statistically estimating whatever hidden factor reliably co-varies with observable performance on those tasks \cite{spearman_general_1904, gould_mismeasure_1981, carroll_1993, deary_125_2012, warne_spearmans_2019}. Spearman's \emph{g} is that hidden factor expressing ``shared variance across a set of intercorrelating cognitive tasks'' \cite{warne_spearmans_2019}.\footnote{Current research attributes as much as nearly half of the variance in IQ scores to \emph{g} \cite{deary_intelligence_2012}.}

Is reliance on \emph{g} enough to eliminate reliance on ethical, social, and political values in determining the boundaries and objects of human intelligence research? %However, in the case of intelligence research, technical procedure arguably does not suffice to eliminate the place of ethical, social, and political values in decisions about the boundaries of the objects of research \cite{anderson_situated_2002, alexandrova_democratising_2022}.

\subsubsection{Quantitative definitions do not solve the boundary problem}

First, the strategy of relying on \emph{g} at best helps with only one of the many phenomena that interest intelligence researchers. A popular taxonomy in the field, introduced by \citet{carroll_1993}, distinguishes three levels at which variations in performance on specific tasks occur \cite{carroll_1993, deary_intelligence_2012}. Differences in performance on one mental task can correlate with: 
\begin{enumerate}
    \item Variations in general performance on all mental tasks (the level of Spearman's \emph{g}).
    \item Variations in performance on a specific family of mental tasks, in a domain of cognitive functions (e.g. working memory).
    \item Variations in performance on the specific task at hand (e.g. ``digit span'' - listen to and repeat this sequence of numbers).
\end{enumerate}

On its own, \emph{g} does not address the boundary problems for the less general levels of this taxonomy: levels 2 and 3. As Deary notes concerning level 2, researchers disagree about the ``nature of the domains—they can vary in number, name and content between samples depending on the battery applied—and there have long been worries about whether the nature of \emph{g} might vary between cognitive batteries'' \cite{deary_intelligence_2012}. These two less general levels are especially relevant for comparison with ML benchmarks: as we will argue in Section \ref{task-choice}, task selection for ML benchmarks is value-laden in similar ways.

\subsubsection{Boundaries and practical consequences influence each other}

Technical considerations alone cannot resolve questions about the \emph{significance} of the research. Judgments of significance in turn influence how we choose to define intelligence. The first IQ test, developed by Binet and Simon in 1905, was intended to help institutions identify students with learning difficulties, for the purpose of separating them from students of ``normal'' intelligence \cite{carroll_1993, deary_intelligence_2020, mitchell_innate_2018}. Other direct practical applications include labor and healthcare \cite{deary_intelligence_2020}. 

For much of the 20th century, intelligence tests saw practical use in education. For instance, in the UK, they were used for sorting students into ``longer and more-academic'' and ``shorter and less-academic streams of secondary education'' \cite{deary_intelligence_2020}. Although IQ has fallen out of favor for that specific purpose, close analogues remain in use under different names, such as the ``cognitive ability test'' (CAT) in the UK, the results of which are highly correlated with Spearman's \emph{g} \cite{deary_education_2007, deary_intelligence_2020}. Likewise, cognitive tests are used to help detect the onset of dementia \cite{cervilla_premorbid_2004}. When deployed in such practical use cases, the boundaries of the concept of the cognitive ability being tested are in part informed by the use case.

%Thus, the values we attach to potential applications of a scientific concept influence how we define the boundaries of that concept.

These practical consequences mean that choices made about the boundaries and significance of research have ethical ramifications. Choices made in defining intelligence research concepts can thus be value-laden in how they make certain ethical outcomes more likely than others. Similarly, in Section \ref{task-choice}, we argue that choices about what tasks in ML are ``significant'' and what ``domains'' ML benchmarks should cover have ethical ramifications.
Just as practical consequences influence boundaries, so do boundaries have practical consequences. The results of intelligence \emph{research}, especially attempts at \emph{explaining} differences in test results, have a long history of practical uses, especially in justifying social hierarchies and structures of oppression \cite{saini_superior_2019, cave_problem_2020, levine_eugenics_2017, deary_intelligence_2012, anderson_situated_2002}.\footnote{\citet{cave_problem_2020} helpfully explores the lessons of this family of issues around the practical uses of IQ---including in legitimating forms of domination---for the AI space in general. \citet{anderson_situated_2002} points out an important historical twist to the role of intelligence research in justifying social hierarchies. Following \citet{gould_mismeasure_1981}, Anderson argues that a case can be made that the historical debate between Spearman's \emph{g} and its detractors (including Thurstone) was in part motivated by disagreement about whether there is a single social hierarchy of abilities.} For example, disability and neurodiversity advocates have longed pointed out the ableist consequences of standardized testing of cognitive abilities \cite{nair2023critical, parekh2022ableism, pellicano2022annual, sequenzia_intelligence_2018}. \footnote{On the broader topic of intelligence, disability, and race, see also \citet{carlson_intelligence_2017}. On the topic of eugenics and the oppression of people with intellectual disabilities, see \cite{mcconnell_devolution_2022}.}.

Just as importantly, as we see next (Section \ref{validity}), intelligence researchers themselves \emph{do and should} explicitly draw on ethical values in examining the validity of constructs in their research. This makes rejecting the place of ethical values in dealing with questions of boundaries especially implausible.

\subsection{Validity and Values}
\label{validity}

Another important area where values play a role in intelligence research is in questions of validity \cite{anderson_situated_2002}. How can we tell whether a given test, or a given statistical construct (e.g. \emph{g}), measures something real, meaningful, or useful? Internal validity, which deals with the internal self-consistency of a concept, maps on to the idea of cross-validation accuracy in ML---the idea that a valid measurement of a construct should provide similar outcomes when applied to similar distributions of data. However, social scientists also define different types of \emph{external} validity---pertaining to how far a construct relates to phenomena outside its internal definition \cite{reiss_against_2019}. Here are a few examples.\footnote{See \cite{jacobs_measurement_2021} for a more comprehensive overview. See \cite{wechsler_wechsler_2008} for a helpful example of how recent IQ tests tackle issues of validity.} %\textbf{Content validity}: How theoretically coherent is the construct being measured? Does operationalization of the construct agree with that theoretical understanding? \textbf{Convergent validity}: Does the proposed measurement of a construct agree with other accepted measurements of the construct? \textbf{Hypothesis validity}: How far are the measurements of the construct able to support substantively interesting hypotheses about the construct? \textbf{Predictive validity}: Does the measure agree with properties that are coarsely related to our construct? For example, if we're measuring IQ, does that correlate in expected ways with other properties associated with intelligence?
\begin{itemize}
    \item (\emph{Content validity}) How theoretically coherent is the construct being measured? Do operationalizations of the construct agree with that theoretical understanding?
   \item (\emph{Convergent validity}) Does the proposed measurement of a construct agree with other accepted measurements of the construct?
  \item (\emph{Hypothesis validity}) How far are the measurements of the construct able to support substantively interesting hypotheses about the construct?
    \item (\emph{Predictive validity}) Does the measure agree with properties that are coarsely related to our construct? For example, if we're measuring IQ, does that correlate in expected ways with other properties and outcomes associated with intelligence?
    \item (\emph{Ecological validity}) Does performance on the measure generalize to real-world contexts outside of the lab?
\end{itemize}
These types of validity and others have been considered when evaluating human intelligence measures \cite{wechsler_wechsler_2008}.

Even if we restrict our attention just to predictive validity, we still face choices of which types of predictive validity to favor. In \citet{jensen_straight_1981}'s terms, IQ tests have validity if they ``improve prediction of the quality of a person’s performance in a larger, more important sphere of activity''; if they can predict outcomes ``that people deem important'' \cite{anderson_situated_2002}. This is a common strategy for establishing that mental tests reliably relate to outcomes that are independently meaningful turns on ethical values: health, educational, and occupational outcomes matter because people value them \footnote{In discussing the validity of IQ, intelligence researchers frequently emphasize that: (a) IQ predicts health and longevity outcomes \cite{gottfredson_why_1997, jensen_g_1998, gottfredson_intelligence_2004, warne_five_2016, arden_association_2016, warne_between-group_2021, warne_between-group_2021}; (b) IQ predicts \emph{education achievement} \cite{deary_education_2007, ritchie_intelligence_2015}; (c) IQ predicts \emph{occupational achievement}\cite{deary_intergenerational_2005, strenze_intelligence_2007, ritchie_intelligence_2015}.}. 

Looking beyond predictive validity, one must also decide which of the other types of external validity described above to prioritize. It is unclear how to make trade-offs between the different types of validity without reference to which practical goals matter more than others. Determining which types of validity are required to accept a benchmark like IQ is a value-laden choice. It depends on how much we care about the external concepts that IQ is purportedly related to.

We believe that taking into account ethical considerations in choosing standards of evidence is something that intelligence researchers \emph{ought to do}, rather than a mere mistake. We see this as an instance of what is sometimes called the \emph{argument from inductive risk}: when the social costs of errors are especially high, ethical, social, and political values \emph{should} influence scientific standards of evidence \cite{douglas_inductive_2000, steel_epistemic_2010}.

\subsection{Path dependence}
\label{self}
We want to consider one last respect in which intelligence research is value-laden: through \emph{path dependence}---mechanisms that ``lock in'' historical antecedents and raise the switching cost of their alternatives \cite{liebowitz1995path, peacock_path_2009, leif_2022}. To paraphrase Anderson, IQ has different ways of becoming a ``self-fulfilling prophecy'' \cite{anderson_situated_2002}.

One aspect of IQ and intelligence research that exhibits path dependence has to do with the real-world outcomes of intelligence research and IQ tests. Intelligence research and IQ tests influence behavior. An especially important area of influence concerns how historically marginalized groups perform on these tests. Take the case of the so-called ``black-white IQ gap'' in the US \cite{jencks_blackwhite_1998, anderson_situated_2002}. Antiracist researchers have examined the role of mechanisms like \emph{teacher expectation} and \emph{stereotype threat} in reinforcing and entrenching historical disparities in cross-group test scores.\footnote{Here, we focus on race. For an overview of research on gender disparities in IQ scores, see \citet{deary_125_2012}.} For instance, false teacher beliefs about the lower potential of black students might lead students to ``discouragement and disengagement from academic achievement'' \cite{ferguson_ronald_teachers_1998, anderson_situated_2002}. 
If IQ tests are used to sort students into groups that receive different resources, those resources can influence how students do on later cognitive tests, thus making it seem like IQ \emph{explains} those real-world outcomes, even if it was the differential allocation of resources or stereotypes that was really responsible. Using IQ tests in this way creates a positive feedback loop that fulfills the ``prophecy'' that IQ is explanatory---even if it isn't.

Human intelligence research today may not look anything like the alternatives that antiracist and feminist critiques of the field have proposed. But intelligence research doesn't necessarily have to reinforce racism and sexism: antiracist and feminist researchers have shown how we could use research to discover mechanisms that reinforce path dependence, and to examine how such mechanisms can be countered and resisted. Whether or not to focus intelligence research in directions that reinforce or help undermine inequities is a value-laden choice.

\section{Values in ML Benchmarks}
\label{ml-values}

Having established that the debates over the correct measures for human intelligence were unavoidably value-laden, we now investigate the relevance of these insights to measures of artificial intelligence. Just as the concept of (human) intelligence is a thick evaluative concept (see Section \ref{human-thick}), the concept of artificial intelligence (in the gradable sense of how intelligent a computer system is) is also a thick evaluative concept.\footnote{Since AI systems often aim to mimic human cognitive abilities, this should be no surprise.} 

The most direct parallels between human intelligence research and ML benchmarks lie in their respective attempts at setting standards for describing, evaluating, and comparing how different models or persons perform on tasks.\footnote{An important difference between ML benchmarks and human intelligence difference is the latter's interest in the environmental and hereditary causes of differences in intelligence. See \ref{crosscultural} for our brief overview of the human intelligence case, and \citet{deary_intelligence_2012} for a more detailed overview of different causal questions in intelligence research.}
ML benchmarks need to enable comparisons of model performance in ways that involve both description and evaluation through:
\begin{enumerate}
    \item  Conveying commensurable empirical facts about the performance of different systems in a way that enables comparisons of model performance; and
    \item  Conveying commensurable content about how we evaluate the systems (approval or disapproval, praise or blame, success or failure, etc.), in a way that enables the ranking of model performance.
\end{enumerate}

Evaluations of ML systems fulfill criterion 1: when researchers evaluate AI systems, they take themselves to be discovering empirical facts about how systems of a certain type behave. They also satisfy criterion 2: when we evaluate an ML system as being better or worse on a certain benchmark, we are evaluating it as being more or less successful. Given that ML evaluation has both descriptive and evaluative components, we can expect similar value-laden issues around ML evaluation to emerge, parallel to the issues we discussed with respect to measuring human intelligence. We now argue for three specific ways in which ethical values and risks can enter ML benchmarks: in task choice (or task weightings), in selecting standards of validity that benchmarks should fulfill, and in creating path dependence through  positive feedback loops in the ML ecosystem of model architectures, hardware, software, human incentives, and benchmarks.

\subsection{Task choices, task scopes, and task weightings are value-laden}
\label{task-choice}

One of the many value-laden choices in benchmark design is what specific tasks to include. For instance, unlike the popular natural language processing (NLP) benchmarks GLUE and SuperGLUE, the new Holistic Evaluation of Language Models (HELM) benchmark \cite{liang_holistic_2022} includes a toxicity detection task as part of its gauntlet of tasks for evaluating language models. What is especially interesting for our purposes is the shift in the conception of task selection between benchmarks like SuperGLUE and HELM. Unlike SuperGLUE, HELM explicitly and reflectively frames task selection as a value-laden choice. 

As examined by \citet{raji2021ai}, benchmarks that attempt to measure more general capabilities, such as ``general language understanding'', have become more popular. The authors of SuperGLUE \cite{wang_superglue_2019}, a successor to GLUE \cite{wang2018glue}, describe the motivation of the benchmark as ``to provide a simple, hard-to-game measure of progress toward general-purpose language understanding technologies for English.'' Toward this goal, the benchmark is designed to incorporate a group of 8 more specific language tasks. The benchmark also provides a single score (e.g. for use in a leaderboard) ``by averaging scores of all tasks.'' \footnote{See also \citet{dehghani_benchmark_2021} on the related trend of meta benchmarks that provide aggregate scores on multiple different tasks. Besides GLUE and SuperGLUE, other examples include XTREME \cite{hu_xtreme_2020}, VTAB \cite{zhai_VTAB_2019}, and RL Unplugged \cite{gulcehre_rl_unplugged_2020}.}

\citet{raji2021ai} argue that the idea of a ``universal'' benchmark that is appropriate for evaluating AI across all contexts is unattainable and unhelpful. Their point is analogous to debates in measuring human intelligence about whether there is a single general factor that explains variations in test scores, and if so, how it should be construed (see Section \ref{crosscultural}). Debates over which tasks to include in cognitive tests for human intelligence parallel debates over which tasks to include in ML benchmarks. In both cases, the selection or weighting of tasks is made with reference to what we want our measurements to do outside of the research environment. What abilities do we want to incentivize AI systems (or humans, in the IQ case) to have? What abilities do we think are not worth incentivizing? These are ultimately questions that cannot be answered without referring to our ethical values around what types of intelligent systems we consider socially good and worth pursuing.

If we accept that there is no such thing as a ``general'' task on which all AI systems should be evaluated, then the question arises of which tasks we \emph{should} use to evaluate AI. The space of possible tasks is at the very least extremely large and possibly infinite. It is not feasible to evaluate AI on \emph{all} tasks. In the case of benchmarks like SuperGLUE, \citet{raji2021ai} argue that task selection is arbitrary and unsystematic: appearing motivated especially by ``convenience'' (such as by what is ``easily available'' and front of mind for those building the benchmark) and largely disconnected from domain knowledge about the general capability being purportedly benchmarked---e.g. of natural language understanding. 

These points parallel Anderson's argument that even if we grant that there are value-neutral scientific facts, the choice of which of these facts are significant enough to seek out is value-laden \cite{anderson_situated_2002}.
Given limited scientific resources, we cannot possibly seek out all undiscovered facts out there. Choices about which truths and facts to pursue should be partly informed by their ethical, social, and political implications. For example, we should start asking questions like why a task like object recognition is prioritized in computer vision (CV) benchmarks. While there is a \emph{prima facie} scientific reason for including object recognition as a task, that is still only one part of human vision, which includes many other abilities beyond object recognition. One potential downside of having object recognition as the paradigm CV task is that it puts the question of curating a set of ``appropriate'' image labels front-and-center. This leads to tensions when CV is applied to situations where the act of labeling itself is contentious and ethically laden: for example, when applied to automatic gender detection \cite{scheuerman-gender, scheuerman-identity}. Furthermore, social scientists have long recognized that the act of classification itself is moral and political \cite{bowker2000sorting}.

HELM \cite{liang_holistic_2022} explicitly embraces the value-laden character of benchmarks, and deliberately articulates a strategy for navigating what this implies in its approach to task selection. Rather than selecting tasks that are a proxy for putative progress towards a more general ability, HELM targets an evolving and revisable set of tasks illustrative of important and socially significant real-world use cases for large language models (``scenarios''), accompanied by a pluralistic set of metrics for measuring performance on those many scenarios. In recognizing the ethical and societal considerations that motivate their selection of tasks and metrics, the authors also explicitly acknowledge: 
\begin{enumerate}
    \item  That tasks and metrics are ``complex and contested social constructs''.
    \item  The importance of designing the benchmark in a way that helps better articulate the ``explicit potential trade-offs'' between the ``desiderata'' these constructs embody, including  ``help[ing] to ensure these desiderata are not treated as second-class citizens to accuracy''.
    \item  That the task of model evaluation should be about coming up with a holistic and plural set of desiderata that helps examine whether the models evaluated are ``societally beneficial systems''.
    \item  The importance of articulating the incompleteness of the task selection, and of foregrounding limitations that call for future work, alternative benchmarks, or revisions to the benchmark. HELM is a ``living benchmark'' designed to ``evolve according to the technology, applications, and social concerns''.
\end{enumerate}

In examining the limitations of their benchmark design, HELM highlights another value-laden trade-off in task selection: between choosing tasks and metrics that enable straightforward ranking of models (e.g. through leaderboards built with ``single-number metrics'') and choosing tasks and metrics that make ranking model performance in a decontextualized way more difficult. As the authors highlight, in some circumstances, benchmarks that enable straightforward rankings or aggregate scores can be appropriate: such as to ``simplify decision-making''. \cite{liang_holistic_2022} However, this is itself a choice always motivated by social, ethical, and political preferences. There is no value-neutral path to benchmarks with leaderboards. 

\subsection{Choices of validity in ML}
\label{ml-validity}

In Section \ref{validity}, we discussed how choices about which types of construct validity to use are value-laden. Here, we argue that similar concerns apply to ML benchmarks.  

Typically, when ML researchers use cross-validation accuracy as the main metric of performance, they are measuring only internal validity, since the examples being validated on are randomly selected from the same dataset that the training data came from \cite{liao_are_2021}. In contrast, the types of validity described in Section \ref{validity} are forms of external validity---relating the findings of a study to external phenomena outside of the study's dataset. In benchmarks that test only for internal validity, choosing not to consider external validity is a value-laden choice.

Some ML methods, like one-shot learning, are inherently evaluated in ways that go beyond internal validity, because they are evaluated on data distributions that are different from the training distribution. However, this is just \emph{one} type of external validity---ecological validity (see Section \ref{validity}). Even when ML researchers do decide to measure external validity, there are further value-laden choices to be made about which types of external validity to measure. In Section \ref{validity} we discussed how external validity can be about the content of what's measured (content validity), the convergence of the measurement with other ways of measuring the same construct (convergent validity), and whether the measurement correlates in the expected ways with other phenomena (predictive validity). Furthermore, even \emph{within} predictive validity, there are different targets that one can select as the phenomenon to find correlations with. Just as human intelligence researchers had to decide if IQ or other cognitive tests should predict health, education, or occupational achievements (among many other choices), ML researchers have to decide what types of external tasks or environments they want their models to best succeed at. In other words, given that an ML researcher is looking for their model to do well in environments or tasks outside the training data distribution, there are still further choices to be made about \emph{which} environments or tasks these should be.

Following \citet{raji2021ai}'s argument that there is no such thing as an ``everything in the whole wide world'' benchmark, it is not feasible for models to correlate equally well with \emph{everything} in the external world. The choices about which external phenomena we want the results of models to best correlate with inevitably depend on how we differentially value different external phenomena. For example, how important is it to design benchmarks that are not biased towards American English \cite{johnson_ghost_2022}? Is it even possible to pay \emph{equal} attention to every language or culture when designing benchmarks? What counts as a ``natural'' photo or video for the purposes of a benchmark dataset (what external phenomena count as ``natural'') \cite{scheuerman2021datasets}? How similar should machine vision be to human vision? What novel physical environments do we want a robot to be able to navigate? For toxicity detection benchmarks in NLP, should the labels in the benchmark dataset account for cross-geographic differences, or is it good enough to model toxicity in a specific culture \cite{ghosh_detecting_2021}? For benchmarks with annotated datasets, the external validity of model performance on the benchmark may depend on how disagreements between annotators are handled when creating the benchmark \cite{aroyo_truth_2015, diaz_crowdworksheets_2022}. \citet{liao_are_2021} emphasize the challenges in ML with establishing whether ``progress on a benchmark transfers to other problems''. We argue here that the choice of \emph{which} problems to measure these transfers on is inevitably value-laden, because it depends on what we want models to do in the real world, which in turn depends on our values.

Finally, failing to carefully consider external validity or selecting an inappropriate from of external validity can have direct ethical consequences. Many high-profile ML failures can be construed as failures of external validity \cite{kelly2019key, moderation, obermeyer}. Many critiques of ML attempts to ``predict'' properties like ``criminality'' or emotions can be interpreted as criticisms of the external validity of these constructs---for example, that criminality is a construct whose content cannot be captured in the way some ML practitioners attempt to (i.e. it lacks content validity) \cite{coston-validity}, or that emotions are defined in a psychologically implausible way by emotion detection algorithms (violating content validity and likely convergent validity as well) \cite{stark2021ethics}.

\subsection{Path dependence}
\label{path-dependency}

Just as measures of human intelligence can become self-fulfilling prophecies (Section \ref{self}),
we argue that similar self-fulfilling dynamics can occur in AI evaluation.  Intelligence tests for humans can be used to place certain humans in lower-resource environments that then cause them to do poorly on future intelligence tests. Similarly, ML benchmarks can discourage work on certain types of models, which in turn causes these types of models to do poorly (according to similar benchmarks) in the future. Current ML benchmarks are dominated by certain types of models---to wit, transformers and other types of deep neural networks \cite{sota2022imagenet, sota2022object, sota2022segment, sota2022translation, sota2022speech}. Some have argued that the availability of certain types of data, the exigencies of current hardware, and the cultural prominence of certain types of problems among AI developers are at least partially responsible for this \cite{hooker2021hardware, ensmenger2012chess, bisk-etal-2020-experience, dotan_value-laden_2020}. For example, \citet{ensmenger2012chess} has argued that the preeminence of chess-playing as a paradigmatic ML problem has influenced how ML developed, similarly to how the practice of using \emph{Drosophila} as a model organism in biology influenced the paths that 20th-century biology took.

\subsubsection{Lotteries}

Path dependence can reinforce itself. Sara Hooker examines hardware-software interdependence as another source of self-reinforcing dynamics: some types of hardware, suitable to certain types of machine learning models or problems, have a healthy software ecosystem that can run on them. Other less dominant types of hardware tend to have less developed software ecosystems associated with them. This puts the latter at an increasing disadvantage \cite{hooker2021hardware}. Hooker identifies multiple instances in computer science history of when  ``a research idea wins because it is suited to the available software and hardware and not because the idea is superior to alternative research directions.''

\citet{dehghani_benchmark_2021} provide an empirical examination of similar feedback loops between benchmarks and models. They identify four problematic dynamics around ML benchmarks:
\begin{enumerate}
    \item Task selection bias: the dependence of model performance on the tasks and datasets selected for the benchmark.
    \item Community bias: the effects of community pressures on the incentives of researchers in choosing benchmarks.
    \item Statefulness of benchmarking: where decisions made in developing new models are penalized or rewarded for the extent to which they are informed by the errors and successes of previous models on the same benchmark. 
    \item Rigging: for model families where community agreement on benchmarks and evaluation best practices is lacking, researchers are incentivized to select evaluation methods (datasets and metrics) that best fit their model and resource constraints. 
\end{enumerate}
 
These dynamics can lead to positive feedback loops that entrench the dominance of certain types of models. As long as performance on currently popular benchmarks is what directs investment in modeling approaches that do well on those benchmarks, other modeling approaches will be increasingly at a disadvantage as they suffer from a relative lack of computational resources, hardware incompatibilities, and less researcher interest \cite{leif_2022}.
For example, we do not know if Bayesian models could achieve similar performances on the same problems if we were to pour the same amount of resources into them that we do for neural networks: resources to improve their architectures, fine-tune them, create software or hardware for them, and train them on large amounts of data. Compared to some other modeling approaches, neural networks get comparatively more investment in the creation and maintenance of sophisticated, open-source software packages like Tensorflow and Pytorch. Relying on benchmarks that largely favor neural networks could give this feedback loop an additional boost.

The much-criticized practice of ``SOTA chasing'' (creating models with the main purpose of getting the best ``state-of-the-art'' performances on benchmarks) \cite{church2022emerging} can be construed as a form of path dependence. It encourages investment into incremental improvements on dominant architectures and modeling approaches. In a different world, we might instead have more investment in less popular model types that are harder to work with under the current software/hardware environment.  

In a different world, one can also imagine having an evaluation environment that corrects for some of the disadvantages of non-dominant model types. For example, some NLP researchers have called for performance measures that adjust for the extent of hyperparameter tuning and computational budget \cite{dodge-etal-2019-show}.
The fact that doing so is not currently widespread practice is a value-laden choice that puts non-dominant model architectures at a disadvantage. However, even if we did correct for computational resources and fine-tuning, we cannot correct for the problem that far more researcher-hours are spent improving neural networks and exploring their possibilities, with comparatively much less effort being spent on other alternatives. As a corrective, research funds could be distributed with more attention to the need to not over-focus on already-popular approaches---but whether to do this or not is a value-laden choice that, given current uncertainties about which path is best, cannot be decided by purely technical considerations.

We also have some potential examples of path dependence in computer vision. Take the overwhelming emphasis in computer vision on objection recognition tasks. These tasks allow computers to score well without necessarily using the same features that humans would use to identify the same objects. For example, it is well-established that computers use texture much more than humans do for object recognition, while using shape much less than humans do \cite{NEURIPS2021_c8877cff}. Current benchmark tasks in computer vision are largely designed such that models are not penalized for this predilection. What would our models look like if we wanted them to more closely mimic humans' greater reliance on shape relative to texture? How would we design benchmarks if this was a priority?\footnote{This is not to say that it \emph{should} be a priority, but it could be beneficial to have more of a debate about whether it should.}

A related issue in computer vision is the question of robustness. Adversarial examples show that popular computer vision models are sensitive to changes in images that are not perceptible to humans \cite{NEURIPS2021_c8877cff}.
The fact that this is common in models that perform well on popular CV benchmarks raises the question of why robustness isn't included as a metric in those benchmarks. What consequences (of non-robust models being deployed in real applications) are we implicitly sanctioning when we leave robustness out of benchmarks? What types of model architectures are we implicitly encouraging?

The increasing dominance of sequence models in NLP is another area of potential path dependence. As \citet{dehghani_benchmark_2021} point out, 7 out of 8 tasks in GLUE are sequence-matching tasks, even though it is not obvious if this is the best composition of tasks to capture natural language understanding. How much did this task composition play a role in elevating transformers, which are particularly adept at extracting sequential information? What paths in NLP were not explored or under-explored because candidate model architectures that picked up on other aspects of language that correlate less with next-token prediction did not perform well on benchmarks like GLUE?

Once we acknowledge path dependence as a real phenomenon and recognize that choosing certain benchmarks can make the playing field less level because of how it influences future investment in ML, it becomes crucial to recognize this as an area of value-laden decisions. We frame this as a choice between the following:
\begin{itemize}
    \item  Validating model performance against current benchmarks, assuming current distributions of community pressures and resources for the development of different types of models.
    \item  Validating model performance in a hypothetical domain where all model types have been tuned to the same extent, use similar computational resources, and have received similar investments in architectural improvements and software ecosystems.
\end{itemize}
We do not intend here to insist that one or the other choice is the \emph{correct} one, but to point out that choosing one or the other is not merely a matter of fact or straightforwardly dictated by the ``scientific method''. Ethical, social and political values ought to inform our choices here, because the homogenization of the model landscape has ethical ramifications and risks \cite{leif_2022, Kleinberge2018340118, creel2022algorithmic, bommasani_picking_2022}.

\subsubsection{Path dependence in reifying social constructs}

Another form of path dependency is when social constructs used as categories in datasets are themselves made ``more real'' or more persistent by the fact that they are being used in benchmarks. In Section \ref{self}, we explained how using IQ tests to differentially allocate resources to students can make IQ itself seem more ``real'' and significant. The differential allocation of resources based on IQ can lead to systematic differences in students' later achievements in life, thus giving IQ a semblance of external validity. Similarly, the increasing use of ML to, for example, classify images of humans into gender categories can reinforce the very same visual norms about gender categories that the ML models presume to be objective \cite{scheuerman-gender}. Using benchmarks that contain such categories encourages ML practitioners to develop models that do well on those benchmarks---which  means the models have to do well at mimicking the categories presumed in the benchmark's ``ground truth'' dataset. These models can then be used in the wider world to classify real humans into gender categories, which means that real humans may be motivated to conform to the visual norms of those categories in order to be classified ``correctly'' by the models. When real humans start modifying their behavior in this way, the norms of gendered visual presentations are strengthened, meaning that apparent gender categories in real-world images that future models will be trained on are also accentuated.

Thus, ML evaluation runs the risk of reifying various social categories or norms that could otherwise be less rigid. Deciding whether or not to include those types of categorizations in our benchmarks is therefore a value-laden choice that depends on whether we think those are good categories to reify. Previous work has criticized current practices around gender categories in ML \cite{scheuerman-gender}.
Many other social categories are also value-laden in similar ways. Although the creators of benchmarks like ImageNet might think that they are simply doing a value-neutral catalog of images of ``the world'' \cite{raji2021ai}, their choices of image labels are deeply value-laden, as others have argued \cite{denton2021genealogy, scheuerman-identity}.

\section{Practical recommendations}
\label{practical}

We have argued that in at least three ways, scientific and technical decisions about ML benchmarks are dependent on ethical values. From the perspective of identifying and mitigating the ethical risks of ML benchmark research, the areas of task selection, choice of validity standards, and path dependence each need close and thorough scrutiny. Being reflective, explicit, and public about the social, political, and ethical values behind ML research is vital to the pursuit of responsible ML.\footnote{On the importance of reflexivity, of critically examining one's own assumptions, in mitigating ML failures, see \cite{boyarskaya_overcoming_2020}. See also \citet{prunkl_institutionalizing_2021}.} Here we outline some practical recommendations for identifying and ameliorating the risks we've identified.

\subsection{Explicitly discussing and documenting ethical risks and values}

A first recommendation is aimed at researchers. Given the unique role of benchmarks in ML research, researchers should avoid interpreting benchmark results as a value-neutral indicator of intelligence or performance. Moreover, we recommend that researchers developing benchmarks explicitly and reflectively discuss the potential ethical risks that come with areas like task selection, standards of validity, and path dependence, either through part of benchmark documentation, as part of impact statements, or as part of the papers themselves. This will in turn help other practitioners choose which benchmarks to use when evaluating their models. For example, if a particular form of external validity (e.g. consequential validity) is particularly important to a real-world application, practitioners who are evaluating models can more easily choose benchmarks that emphasize consequential validity, while putting less weight on benchmarks that don’t. This will better align model evaluation with the values the practitioner wants. 

Building benchmarks in a way that meaningfully centers their value-laden features is not an unachievable ideal. As discussed above (Section \ref{task-choice}), we see HELM \cite{liang_holistic_2022} as a helpful example of  how this can be done for the social, political and ethical values involved in task selection. Notably, HELM repeatedly and constructively draws on recent position papers critical of ML benchmarking practices, including \cite{raji2021ai, liao_are_2021, bowman_what_2021}. Critical position papers have a vital role to play in improving benchmarking practices. They must be given a prominent place in conferences and publications aimed at the ML benchmark community.

\subsection{Ethics reviews and guidelines}

A second recommendation is for people working on ethics guidelines and ethics reviews for ML benchmarks in communities like NeurIPS's Datasets and Benchmarks track. Since 2020, the NeurIPS conference has been systematizing its approach to taking ethical risks into account \cite{prunkl_institutionalizing_2021, srikumar_advancing_2022}. As part of this work, the conference made public new ethical review guidelines in 2021 \cite{luccioni_ethical_2021}. In 2022, a Provisional Draft of the NeurIPS Code of Ethics was published \cite{bengio_provisional_2022}, addressing topics like the uses of research and the treatment of human subjects \cite{lovelace_ethics_2022, lovelace_looking_2022}. A final version of the Code of Ethics was released in 2023 \cite{samy_bengio_announcing_2023}. Parallel work on ethics review is happening in journals, such as Nature Machine Intelligence \cite{srikumar_advancing_2022, editorial_how_2021}. Finally, there is a growing body of work examining the ML ethics review in the context of institutions: from institutional review boards (IRB) and research ethics committees (REC) within research institutions, to impact requirements by funding bodies \cite{prunkl_institutionalizing_2021, lovelace_looking_2022, srikumar_advancing_2022}. For simplicity, we focus on the case of ethics review as part of the research peer review process. But the issues we highlight have parallels in contexts like IRB and funding bodies.  
Given the unique role of ML benchmarks in enabling the evaluation and comparison of ML models, examinations of the ethical risks of benchmarks should not be narrowly limited to considerations about their potential uses, or about the treatment of human subjects \cite{metcalf_where_2016, lovelace_looking_2022}. 

We see the problem of improving ethics review for benchmarks in those areas as a very difficult one. The problems posed by the low maturity of the ML ethics review ecosystem \cite{srikumar_advancing_2022} are arguably compounded by the low maturity level of benchmarking practices in those areas. We lack knowledge of how benchmark ethics review affects researcher behavior, of the design space for benchmark ethics review mechanisms, and of how to ensure the legitimacy of benchmark ethics review---such as through public deliberation \cite{srikumar_advancing_2022, prunkl_institutionalizing_2021}. We also lack knowledge of what improving benchmarking practices in any of the three areas we identified can look like. In the area of task-selection, we noted that HELM is an early example of what meaningfully centers the value-laden features of benchmarks might look like. Yet many more such examples are needed to empower benchmark ethics reviewers to rigorously distinguish what behaviors to flag or to promote for the ethical risks of task-selection. In the area of validity, the example of HELM is also instructive. The authors pointedly acknowledge that there is a dire need for future work interrogating the validity of datasets in the context of HELM's tasks and metrics (scenarios) \cite{liang_holistic_2022}. We lack the collective knowledge to empower ethics reviewers to rigorously distinguish what adequate anticipation of the ethical risks of benchmark validity looks like. 

In the short term, here are two low-hanging fruit proposals that ethics review could begin implementing on the three areas of risk we have identified. First, conferences and publications could add guidelines for ethics reviewers that specifically center the ethical risks of task-selection, validity, and path dependence. What does it look like to helpfully invite reviewers to consider those areas of risk? What challenges stand in the way of making such guidelines actionable?

Second, conferences and publications conducting ethics reviews could consider curating lists of accepted papers that do an especially excellent job engaging with the ethical risks of task-selection, validity, and path dependence. This is a similar spirit to \citet{prunkl_institutionalizing_2021}'s suggestion to implement best impact statements prizes for the sake of incentivizing high-quality impact statements. But rather than only focusing on incentives, we would like to see experimentation with the role ethics review can play in curating documentation that supports the benchmark community's efforts in pushing best ethical risk practices forward. Best impact statement prizes stand on the more selective end of that spectrum. We want to invite the community to also consider less selective approaches. 

\subsection{Structural solutions for the overall research landscape}

Finally, we want to acknowledge an important practical challenge in mitigating the ethical risks that come with the areas we have identified. Path dependence is an example of a \emph{structural problem} that is unlikely to be resolved through isolated individual action. Mitigating the ethical risks that come with path dependence calls for social and collective change in areas like how research is funded and incentivized. For the ethical risks of benchmarks that stem from structural problems like path dependence, future work should investigate strategies and opportunities for leveling the playing field for diverse approaches to ML, in order to minimize path dependence. \footnote{\citet{lovelace_looking_2022} helpfully emphasizes the importance and difficulty of tackling the structural, ecosystem-level sources of ethical risk in ML research.}

\section{Conclusion}
\label{good-science}
What does \emph{good science} look like in the context of ML benchmarks? We've argued that benchmarks are and should be influenced by ethical, social, and political values. The concept of intelligence is value-laden: we define what intelligence is according to how we want machines or people to perform. Moreover, decisions about benchmarks for ML and human intelligence are intimately tied to their uses. The practical implications of ML benchmarks and IQ mean that the social costs of errors can be very high. 

The argument from inductive risk (see Section \ref{validity}) applies to each of the dimensions of the problem we considered for both ML benchmark and human intelligence research. Decisions like what tasks we want machines (and humans) to perform well, what standards of validity we want to use for either, and how to manage path dependence \emph{should be} deeply informed by what purposes we want them to serve in society.

However, there is still a role for value-neutrality in science. For scientific research to be valuable, it must produce reliable empirical knowledge \cite{douglas_moral_2014}. This requires prioritizing considerations about what is more or less conducive to the truth---about \emph{epistemic values}---throughout. Key background decisions (about task selection, standards of validity, and whether to reinforce or curb path dependence) must be informed by both epistemic considerations and ethical values. Yet once we have answered these background questions, we should use only epistemic values to evaluate research.\footnote{This is similar to Anderson's take on impartiality \cite{anderson_situated_2002}.} Paraphrasing \citet{douglas_moral_2014}, intelligence is built on a social, political, and moral terrain. Good science for ML benchmarks and human intelligence research requires reflective and explicit articulation of the values and risks embedded in their scientific and technical core.\footnote{For more on the conception of good science we recommend for benchmarks, see \ref{value-neutrality}.} 

We want to conclude by highlighting an ideal for good science that we especially hope to see explored by the ML benchmark community. Once we accept that benchmarking intelligence is value-laden---that ethical, social and political values should inform how we select standards for describing, evaluating, and comparing human or machine intelligence---we can begin to consider how to define benchmarks in order to promote the values that we want. In the case of measuring human intelligence, \citet{anderson_situated_2002} suggests explicitly taking up the \emph{epistemological standpoint of justice}. This is an injunction to: 
\begin{enumerate}
    \item Focus the efforts of research on figuring out what it would take for members of all groups to fully develop their potential---as opposed to today's society, where members of vulnerable and historically marginalized communities are systematically deprived of opportunities to do so.
    \item Thoroughly investigate what obstacles (e.g. path dependency) get in the way. 
\end{enumerate}
It's an epistemological standpoint because our response to the injunction should turn on ``discoverable empirical facts'' and reliable empirical knowledge, rather than merely on non-epistemic values. In the case of human intelligence, examples include studying mechanisms like teacher expectation and stereotype threat (Section \ref{self}), and experimenting with social reforms \cite{anderson_situated_2002}. 

What would it take to adopt the epistemological standpoint of justice for ML benchmarks? For example, what measures of machine intelligence would enable different groups of humans to thrive in a society where they interact regularly with ML systems? What measures of machine intelligence might reinforce harms to vulnerable and historically marginalized communities (e.g. through path dependence and the reification of social constructs)? Misguided belief in ``neutral'' approaches to evaluating intelligence---as though we can simply measure ``everything in the whole wide world'' \cite{raji2021ai}---makes for worse science. 

%%
%% The acknowledgments section is defined using the "acks" environment
%% (and NOT an unnumbered section). This ensures the proper
%% identification of the section in the article metadata, and the
%% consistent spelling of the heading.
\begin{acks}
We are thankful to the NeurIPS 2022 Queer in AI poster session attendees for helpful input on an earlier version of this paper. We are also grateful for thoughtful discussions of the paper on the BABL AI podcast (December 8 2022), and at the inaugural Hugging Face Ethics and Society Q\&A (March 13 2023). 
\end{acks}

%%
%% The next two lines define the bibliography style to be used, and
%% the bibliography file.
\bibliographystyle{ACM-Reference-Format}
\bibliography{paper}

%%
%% If your work has an appendix, this is the place to put it.
\appendix

\section{Appendix}

\subsection{Value neutrality and feminist philosophy of science}
\label{value-neutrality}

Our perspective in this paper is deeply informed by feminist philosophy of science scholarship about the value neutrality of science. In this appendix, we provide background on the conception of value neutrality implicit in this paper (\ref{background-neutrality}). We also argue that feminist critiques of value neutrality are a vital source of insights for ML research, including for articulating the place of ethical, social, and political values in \emph{good science} (\ref{feminist-neutrality-relevance}). We mean this as an invitation for other researchers to dig further into how this body of scholarship can be helpful to the ML research context.

\subsubsection{What is value-neutrality?}
\label{background-neutrality}
Discussions of the value neutrality of science have a long history. \citet{mcmullin_values_1982} interprets Weber as having argued---over a hundred years ago---that "the objectivity of science [...] requires public norms accessible to all, and interpreted by all in the same way" \cite{weber_sinn_1917, weber_meaning_2011}. On this early 20th century view, values that are open to choice and disagreement have no place in science. 

By contrast, for multiple decades, philosophers have recognized a role for values that are open to dispute and choice in scientific reasoning and decisions \cite{kuhn_objectivity_1977, mcmullin_values_1982, longino_cognitive_1996, dotan_value-laden_2020}.\footnote{For a helpful account connecting philosophy of science to values in ML research, see \citet{dotan_value-laden_2020}. On the philosophy of science front, our accounts center different bodies of work: “Kuhn’s paradigms and Lakatos’s research programmes” in their case; feminist scholarship on the value neutrality of science, epistemic/nonepistemic values, thick evaluative concepts in social science, and the argument from inductive risk in ours.} A particularly important concern is the \emph{underdetermination thesis}: scientific theory is underdetermined by empirical evidence \cite{longino_cognitive_1996, mcmullin_values_1982}. Empirical data cannot on its own settle the question of what specific theory is correct. A commonly accepted view is that values have a role to play in closing the gap between empirical data and theory choice. 

This hasn't spelled the end of the value neutrality of science. We can hold on to the ideal of value-neutral science, some argue, if the only values we rely on in closing the gap between empirical data and theory choice are \emph{epistemic values}: that is, considerations about what promotes the attainment of truth \cite{steel_epistemic_2010, longino_cognitive_1996}. \footnote{In some 90's discussions, epistemic values are instead called "cognitive values". Recent discussions favor the term epistemic values to avoid presupposing (or seeming to presuppose) non-cognitivism about ethical, social, and political values.} Epistemically problematic practices like falsifying records or cherry-picking data undermine the core functions that make science worth pursuing, such as the aim of producing ``reliable empirical knowledge'' \cite{douglas_moral_2014}. Acknowledging a role for epistemic values in scientific reasoning is an invitation to clarify specifically what we consider to promote the attainment of truth or of reliable empirical knowledge. Kuhn's influential account highlights 5 key epistemic values: accuracy, simplicity, internal and external consistency, breadth of scope, and fruitfulness \cite{kuhn_objectivity_1977, longino_cognitive_1996}. 

Of note, this philosophy of science sense of \emph{value neutrality} is different from the sense of value-neutrality centered in \citet{birhane_values_2022}'s helpful investigation of the values of ML research. Their account instead centers \citet{winner_autonomous_1978}'s conception of the value-neutrality: the view that technology is value-neutral if it can be put to both beneficial and harmful uses, and if whether it is harmful or not harmful depends on what uses and purposes we choose to put it to. This is the kind of ``neutrality'' that household tools have. A hammer can be used to attach a work of art to a wall; it can also be used as a weapon. 

The philosophy of science sense of value neutrality is instead about what kinds of values are admissible in scientific reasoning. Contemporary versions of the value neutrality of science thesis allow \emph{epistemic values} to supplement empirical evidence in closing the gap between evidence and theory. What the value-neutrality thesis rejects is any place for ethical, social, or political values (non-epistemic values more broadly) in scientific reasoning.

\subsubsection{The relevance of feminist philosophy of science for ML research}
\label{feminist-neutrality-relevance}
There are two aspects of feminist philosophers of science's rejection of value neutrality that we find especially helpful for the ML space.

The first concerns the status of scientific research that purports or appears to be \emph{value-neutral}---that does not make explicit and reflective appeal to ethical, social, and political values. In this paper, we have argued that IQ and ML benchmark research, even in cases where it purports or appears to be value-neutral,  nonetheless embodies ethical, social and/or political values. We see this as deeply aligned with a growing body of scholarship on the values embodied in ML research \cite{dotan_value-laden_2020, leif_2022, birhane_values_2022}. In doing so, we hope to spark interest in examining the many other ways in which insights from the feminist philosophy of science tradition can illuminate the roles of implicit ethical, social, and political values in ML research.

The second concerns the status of research that is explicitly and reflectively informed by ethical, social, and political values, like the insights of anti-racist and feminist IQ research that Elizabeth Anderson examines (discussed above in Section \ref{human-thick}). Inspired by feminist standpoint epistemology's emphasis on the epistemic privilege of the standpoints of oppressed and historically marginalized communities, feminist philosophers of science have been investigating the multiple ways in which explicit and reflective reliance on ethical, social, and political values can make for \emph{better science} \cite{anderson_situated_2002}. As discussed in Section \ref{good-science}, this perspective is compatible with a strong endorsement of the crucial role of empirical evidence in science, the distinct value of science in producing reliable empirical knowledge, and the impartiality of science. 

A lot of research on ethical and just ML deeply aligns with this perspective. We believe that the feminist philosophy of science tradition is an especially valuable source of models for the multiple ways in which \emph{good science} is and should be informed by ethical, social, and political values.

\end{document}